# Performance Evaluation of Machine Learning Classifiers in Sentiment Mining

G.Vinodhini*, RM.Chandrasekaran**
*Assistant Professor, Department of Computer Science and Engineering, Annamalai University,
*Professor,Department of Computer Science and Engineering, Annamalai University,
Annamalai Nagar-608002, India.

*Abstract*

In recent years, the use of machine learning classifiers is of great value in solving a variety of problems in text classification. Sentiment mining is a kind of text classification in which, messages are classified according to sentiment orientation such as positive or negative. This paper extends the idea of evaluating the performance of various classifiers to show their effectiveness in sentiment mining of online product reviews. The product reviews are collected from Amazon reviews. To evaluate the performance of classifiers various evaluation methods like random sampling, linear sampling and bootstrap sampling are used. Our results shows that support vector machine with bootstrap sampling method outperforms others classifiers and sampling methods in terms of misclassification rate.

## I. INTRODUCTION

E-Commerce is becoming an increasingly popular today, where a consumer can buy a product online. After buying a product, consumers can post their reviews and comments about the product in internet. These reviews can be a powerful source of finding out the consumer preferences and making recommendations of products to a new consumer who desires to buy a product. There are many product review sites like Amazon, epinions.com, etc. which provide the reviews for products. If a user who wants to buy a camera, then he goes to a product review site and read the reviews. There could be many thousands of reviews for cameras. It is  practically feasible for the user to read all the reviews. So, a comprehensive system is needed to analyze the reviews and find out the quality of the product. In recent years, the use of machine learning classifiers is of great value in solving a variety of problems in sentiment classification. We focus on sentiment mining which aims to discover reviewers' attitudes, whether positive or negative, of a product.

The major contributions and uniqueness of the work presented in this paper are as follows:
We show the robustness of four of the top classifiers in data mining in terms of their misclassification rate for  sentiment mining.We also analyze the effectiveness of various evaluation methods like random sampling, bootstrap sampling and linear sampling on classifier performance.

The following section presents a brief description of supervised learning followed by with details of one of the top four machine learning classifiers in data mining that are also used in this paper. Section 3 is about the data source used. Section 4 explores the performance of each classifier using the data source. This section also presents results of sampling methods used. Section 5 reviews some related work to the problem of sentiment classification. We close with conclusions and directions for future research.

## II. MACHINE LEARNING ALGORITHMS

In supervised learning, given a set of training instances and any given prior probabilities and misclassification costs, a learning algorithm outputs a classifier. The classifier is an hypothesis about the true classification function that is learned from, or fitted to, training data. The classifier is then tested on test data. A wide range of algorithms in machine learning paradigms have been developed for the task of supervised learning classification. We now discuss the four classifiers used in this work.

### A. Decision Trees

Decision tree induction is one of the simplest forms of supervised learning algorithm. It has been extensively used in many areas such as statistics and machine learning for the purposes of classification and prediction. Decision tree (DT) classifiers can be generalise beyond the training sample so that unseen samples could be classified with as high accuracy as possible. DTs are non-parametric and a useful means of representing the logic embodied in software routines. A decision tree takes as input a case or example described by a set of attribute values, and outputs a Boolean decision [5].  In the classification case, when the response variable takes value in a set of previously defined classes the node is assigned to the class which represents the highest proportion of observations.

### B. K-Nearest Neighbour

One of the most venerable algorithms in machine learning is the nearest neighbour (NN). The entire training set is stored in the memory. To classify a new instance, the Euclidean distance is computed between the instance and each stored training instance and the new instance is assigned the class of the nearest neighbouring instance. More generally, the *k* nearest neighbours are computed, and the new instance is assigned the class that is most frequent among the *k* neighbours [7]. To classify an unknown pattern, the *k*-NN approach looks at a collection of the *k* nearest points and uses a "voting" mechanism to select between them, instead of looking at the single nearest point and classifying according to that with ties broken at random.

### C. Naïve Bayes Classifier

Bayesian learning algorithms use probability theory as an approach to concept classification. Bayesian classifiers produce probabilities for class assignments, rather than a single definite classification. Naïve Bayes classifier (NBC) is perhaps the simplest and most widely studied probabilistic learning method. It learns from the training data, the conditional probability of each attribute $A_i$, given the class label $C$. The strong major assumption is that all attributes $A_i$ are independent given the value of the class $C$. Classification is therefore done applying Bayes rule to compute the probability of $C$ and then predicting the class with the highest posterior probability. The assumption of conditional independence of a collection of random attributes is very critical [8].

### D. Support Vector Machines

Support Vector Machines (SVMs) are pattern classifiers that can be expressed in the form of hyper-planes to discriminate positive instances from negative instances. SVMs have successfully been applied to numerical tasks, including classification. They perform structural risk minimisation and identify key "support vectors". Risk minimisation measures the expected error on an arbitrarily large test set with the given training set and SVMs non-linearly map their *n*-dimensional input space into a high dimensional feature space. In this high dimensional feature space a non-linear classifier is constructed [5][7][8]. Given a set of points which belong to either of two classes, a linear SVM finds the hyper-plane leaving the largest possible fraction of points of the same class on the same side, while maximising the distance of either class from the hyper-plane. The hyper plane is determined by a subset of the points of the two classes, named support vectors, and has a number of interesting theoretical properties [10].

### III. DATA SOURCE

The reviews are collected from www.amazonreviews.com for five different products. The reviews collected are in user free format. So pre-processing of review is needed before classifier is applied. Pre-processing steps, including stop words removal and word stemming, are first applied to the review documents in order to reduce the noisy information in the following processes. These pre-processing steps can improve the performance of sentiment classification. The five different product reviews analysed are camera, mobile, i-pod, laptop and music player. The number of reviews collected for each product varies from 200 to 300. The details of the data used in the analysis are listed in Table 1.

Table 1. Data source

| Product | No.of.review Sentences | Positive reviews | Negative reviews |
|---|---|---|---|
| Camera | 220 | 132 | 88 |
| Mobile | 218 | 146 | 72 |
| i-Pod | 283 | 119 | 164 |
| laptop | 240 | 128 | 112 |
| Music player | 236 | 160 | 76 |

After pre-processing, the reviews are represented as unordered collections of words and the features (Unigram) are modelled as a bag of words. A word vector is created using the unigram features based on the term occurrences.

### IV. EVALUATION

In order to empirically evaluate the performance of the four classifiers, an experiment is used on five different product reviews in terms of misclassification error rate. To perform the experiments, each data is split randomly into ten parts of equal size with 10-fold cross validation used for this task. For each fold, nine parts were placed in the training set and the remaining onewas placed in the corresponding test set . We construct the predictive models using four classifiers from the WEKA toolkit. The WEKA is an ensemble of tools for data classification, regression, clustering, association rules and visualization. The toolkit is developed in Java and is open source software. All the four classifiers are used with their default as implemented in WEKA [5][7].

To measure the performance of these classifiers, the misclassification rate is used. The classifier approaches are cross validated using several types of sampling for building the subsets.The linear sampling simply divides the input dataset into partitions without changing the order of the examples i.e. subsets with consecutive examples are created. The random sampling builds random subsets of the input dataset. Samples are chosen randomly for making subsets.The bootstrap sampling builds random subsets and ensures that the class distribution in the subsets is the same as in the whole input dataset. In the case of a binominal classification, bootstrap sampling builds random subsets such that each subset contains the same proportions of the two values of class labels [2].

Misclassification rate is defined as the ratio of number of wrongly classified review to the total number of reviews classified by the prediction system.

Based on the conducted experiments, the misclassification error rate results for Decision tree, KNN,NB and SVM for various sampling methods are presented in Table 2 for five different products . From the performance measures it is observed that SVM outperforms the other algorithms in terms of overall misclassification rate. KNN is the second better performance than the other classifiers. Among the sampling methods used boot strap sampling performs significantly better than other sampling methods for all classifiers.

V. RELATED WORK

The area of sentiment mining has seen a large increase in academic interest in the last few years. Researchers in the areas of natural language processing, data mining, machine learning, and others have tested a variety of methods of automating the sentiment analysis process.

Pang et al. [4] researched sentiment mining using a binary unigram representation of patterns.In this representation, training patterns are represented by the presence/absence of words instead of by the count of total word occurrences. They tested a variety of algorithms for classification and found that a support vector machine had the highest accuracy of 82.9% using a movie reviews dataset. Whitelaw et al. [11] proposed improving sentiment mining pattern representations by using appraisal groups. They define appraisal groups as "coherent groups of words that express together a particular attitude, such as 'extremely boring', or 'not really very good'." By combining a standard bag-of-words approach with appraisal groups they report a 90.2% classification accuracy. Liu et al [1] proposed a technique based on language pattern mining to identify product features from pros and cons in reviews in the form of short sentences. They also make an effort to extract implicit features. Moreover, Kang et al [3] proposed feature extraction for capturing knowledge from product reviews. In their method, the output of Hu's system was used as the input to their system, and the input was mapped to the user-defined taxonomy features hierarchy thereby eliminating redundancy and providing conceptual organization.

Snyder and Barzilay [9] describe an algorithm that breaks up reviews into multiple aspects and then provides different numerical scores for each aspect. This would be helpful for mixed reviews that explicitly describe those aspects which are good or bad. For example, a movie reviewer may like a movie's acting and special effects, but find its plot poorly conceived. For feature level, it mostly focuses on the extracting and analyzing the product features, finds the commented features, furthermore to generate opinion summaries[5][6][7][8].

VI. CONCLUSION

The major contributions of the paper has been the application of one of the top four machine learning algorithms to predict sentiment orientation of the review sentences and to evaluate various sampling methods on classifier performance .Five different product reviews were utilised for this task. The results suggest that the ML algorithms can be successfully applied in sentiment mining providing significant classification performance. Based on evidence, it has been found that among all classifiers (DT, $k$-NN, NBC and SVM), SVM with bootstrap sampling method performs better in terms of misclassification error rate. While many research continues, practitioners and researchers may apply ensemble machine learning methods for constructing the model to predict sentiment orientation. We plan to replicate our study to predict the models based on ensemble machine learning algorithms and genetic algorithms.

Table 2. Performance of classifiers and sampling methods

| Classifier | Sampling method | Misclassification rate | | | | |
|---|---|---|---|---|---|---|
| | | Camera | Mobile | i-pod | laptop | Music Player |
| Decision Tree | Linear Sampling | 21.2 | 21.5 | 21.1 | 21.1 | 21.8 |
| | Random Sampling | 20.8 | 21.8 | 21.4 | 21.0 | 22.2 |
| | Bootstrap sampling | 20.3 | 20.8 | 19.9 | 20.8 | 20.6 |
| KNN | Linear Sampling | 19.8 | 18.2 | 18.1 | 18.3 | 17.8 |
| | Random Sampling | 18.8 | 18.2 | 17.8 | 18.6 | 17.3 |
| | Bootstrap sampling | 18.5 | 17.7 | 17.3 | 18.0 | 17.1 |
| Naive Bayes | Linear Sampling | 22.2 | 23.8 | 23.1 | 22.1 | 22.8 |
| | Random Sampling | 21.5 | 22.9 | 21.7 | 21.6 | 22.3 |
| | Bootstrap sampling | 21.3 | 21.8 | 20.9 | 20.5 | 20.9 |
| SVM | Linear Sampling | 18.6 | 17.2 | 18.1 | 18.3 | 17.6 |
| | Random Sampling | 18.3 | 17.6 | 17.9 | 18.0 | 17.3 |
| | ***Bootstrap sampling*** | ***16.7*** | ***16.8*** | ***16.6*** | ***16.6*** | ***16.7*** |

REFERENCES

[1]. B. Liu, M. Hu, and J. Cheng, "Opinion Observer: Analyzing and Comparing Opinions on the Web,"Proc. International World Wide Web Conference, Japan, May 2005, pp. 342-351

[2]. Ho, Tin Kam. 1998. The random subspace method for constructing decision forests. IEEE Transactions on Pattern Analysis and Machine Intelligence, 20(8): 832-844.

[3]. Kang Liu, Jun Zhao. "NLPR at Multilingual Opinion Analysis Task in NTCIR7", Proceeding of NTCIT-7 workshop Meeting. 2008. pp:226-231.

[4]. Pang, B., and Lee, L. 2004. A sentimental education: Sentiment analysis using subjectivity summarization based on minimum cuts. In ACL, 271–278.

[5]. Polikar, R. 2006. Ensemble based systems in decision making. IEEE Circuits and Systems Magazine Third Quarter:21–45.

[6]. Popescu,O Etzioni. "Extracting Product Features and Opinions from Reviews," Proceedings of Empirical Methods in Natural Language Processing, 2005. pp.339-346

[7]. Schapire, R. E. 2002. The boosting approach to machine learning: Anoverview.

[8]. Shu Zhang, Wen-Jie Jia, Ying-Ju Xia, Yao Meng, Hao Yu, "Opinion Analysis of Product Reviews", 2009 Sixth International Conference on Fuzzy Systems and Knowledge Discovery, pp.591-595

[9]. Snyder, B., and Barzilay, R. 2007. Multiple aspect ranking using the good grief algorithm. In Proceedings of NAACL HLT, 300–307.

[10]. Tetsuya Nasukawa and Jeonghee Yi, "Sentiment analysis: Capturing favorability using natural language processing," In Proc. of the Second International Conferences on Knowledge Capture, 2003. pp.70-77.

[11]. Whitelaw, C.; Garg, N.; and Argamon, S. 2005. Using appraisal groups for sentiment analysis. In Proceedings of the 2005 ACM CIKM International Conference on Information and Knowledge Management, Bremen,Germany, October 31 - November 5, 2005,625–631. ACM.

.